%% file: main.tex
\documentclass[letterpaper]{article} 
\usepackage[preprint]{aaai2027} 
\usepackage[hyphens]{url} 
\usepackage{graphicx} 
\usepackage{natbib} 
\usepackage{caption} 
\usepackage{amsmath,amsfonts,amssymb}
\usepackage{algorithm}
\usepackage{algorithmic}
\usepackage{multirow}
\usepackage{colortbl}
\usepackage{booktabs}

\providecommand{\IEEEPARstart}[2]{#1#2}
\providecommand{\href}[2]{#2}

\makeatletter
\g@addto@macro\@maketitle{\input{sections/teaser}}
\makeatother

\title{SynAgent: Generalizable Cooperative Humanoid Manipulation via Solo-to-Cooperative Agent Synergy}

\author{
    Wei Yao\equalcontrib\textsuperscript{\rm 1},
    Haohan Ma\equalcontrib\textsuperscript{\rm 2},
    Hongwen Zhang\textsuperscript{\rm 3},
    Liangjun Xing\textsuperscript{\rm 4},\\
    Zhile Yang\textsuperscript{\rm 2},
    Yuanjun Guo\textsuperscript{\rm 2},
    Yunlian Sun\corresponding\textsuperscript{\rm 1},
    Yebin Liu\textsuperscript{\rm 4}
}

\affiliations{
    \textsuperscript{\rm 1}School of Computer Science and Engineering, Nanjing University of Science and Technology, Nanjing 210094, China\\
    \textsuperscript{\rm 2}Shenzhen Institutes of Advanced Technology, Chinese Academy of Sciences, Shenzhen 518055, China\\
    \textsuperscript{\rm 3}School of Artificial Intelligence, Beijing Normal University, Beijing 100875, China\\
    \textsuperscript{\rm 4}Department of Automation, Tsinghua University, Beijing 100084, China\\
    \{wei.yao, yunlian.sun\}@njust.edu.cn,
    \{hh.ma2, zl.yang, yj.guo\}@siat.ac.cn,\\
    zhanghongwen@bnu.edu.cn,
    \{xlj24@mails, liuyebin@mail\}.tsinghua.edu.cn
}

\begin{document}

\maketitle

\input{sections/abstract}
\input{sections/introduction}
\input{sections/related_work}
\input{sections/method}
\input{sections/experiments}
\input{sections/conclusion}

\bibliography{SynAgent}

\clearpage

\appendix
\input{sections/supplementary_material}

\end{document}

%% file: sections/teaser.tex
\par\vspace{0.25em}
\begin{minipage}{\textwidth}
    \centering
    \includegraphics[width=\textwidth]{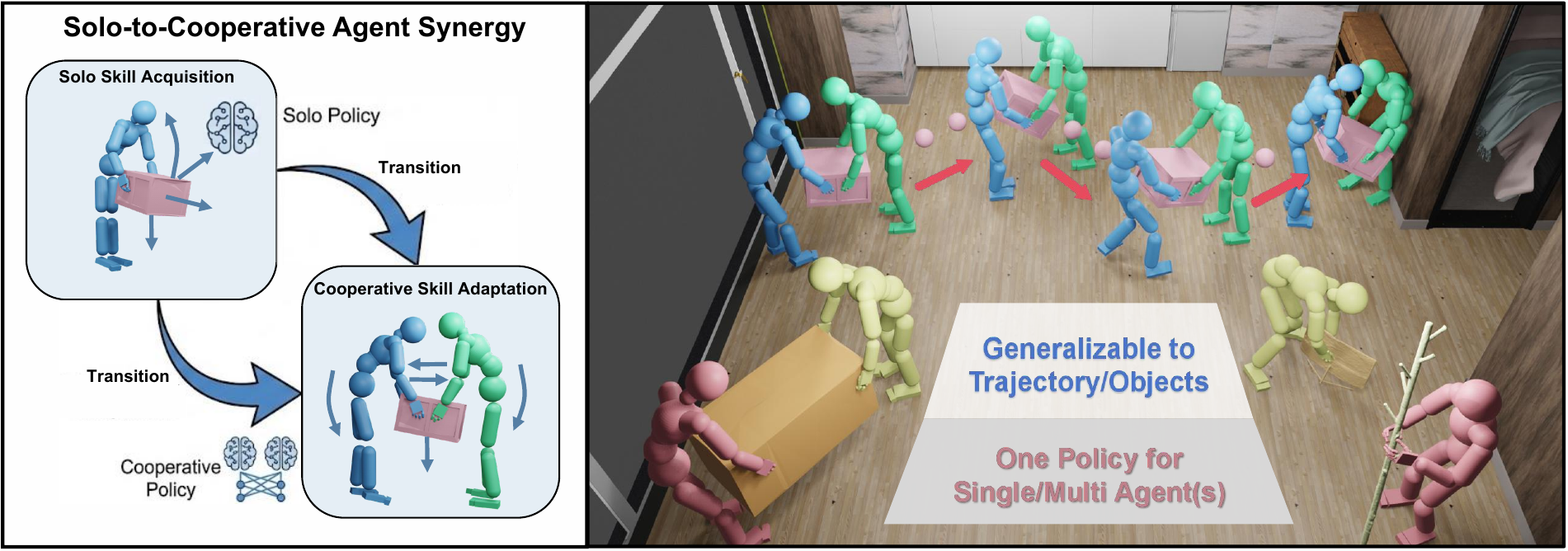}
    \captionof{figure}{\textbf{Features of SynAgent.} As the first model to address trajectory-following object manipulation with multiple humanoid agents, SynAgent generalizes across diverse object geometries and supports cooperative manipulation.}
    \label{fig:teaser}
\end{minipage}
\vspace{0.5em}

%% file: sections/abstract.tex
\begin{abstract}
Controllable cooperative humanoid manipulation is a fundamental yet challenging problem for embodied intelligence, due to severe data scarcity, complexities in multi-agent coordination, and limited generalization across objects. In this paper, we present \textbf{SynAgent}, a unified framework that enables scalable and physically plausible cooperative manipulation by leveraging \textbf{Solo-to-Cooperative Agent Synergy} to transfer skills from single-agent human-object interaction to multi-agent human-object-human scenarios. To maintain semantic integrity during motion transfer, we introduce an interaction-preserving retargeting method based on an Interact Mesh constructed via Delaunay tetrahedralization, which faithfully maintains spatial relationships among humans and objects. Building upon this refined data, we propose a single-agent pretraining and adaptation paradigm that distills synergistic collaborative behaviors from abundant single-human data through decentralized training and multi-agent PPO. Finally, we develop a trajectory-conditioned generative policy using a conditional VAE, trained via multi-teacher distillation from motion imitation priors to achieve stable and controllable object-level trajectory execution. Extensive experiments demonstrate that SynAgent significantly outperforms existing baselines in both cooperative imitation and trajectory-conditioned control, while generalizing across diverse object geometries. Codes and data will be available after publication. Project Page: \textcolor{magenta}{\url{https://yw0208.github.io/synagent/}}.
\end{abstract}

%% file: sections/introduction.tex
\section{Introduction}
\IEEEPARstart{E}{mbodied} intelligence has emerged as a key research frontier due to its substantial potentiality. Most existing work focuses on single-robot locomotion~\cite{luo2023phc, luo2023pulse} and isolated object manipulation~\cite{luo2024omnigrasp, tessler2024maskedmimic}. But future large-scale deployments will require robots to operate synergistically. In such settings, two capabilities are crucial: \textbf{cooperative manipulation} for multi-robot collaboration and \textbf{precise control} for reliable motion execution. Developing these capabilities is essential for advancing from isolated robotic skills to synergistic multi-agent systems.

Despite their importance, robust cooperative manipulation and precise control remain challenging. A key limitation lies in the scarcity of high-quality large-scale datasets that capture human--object--human interactions (HOHI). Most existing datasets focus on single-person motion~\cite{mahmood2019amass}, dual-human interaction~\cite{liang2024intergen, xu2024interx}, and human--object interaction (HOI)~\cite{taheri2020grab, jiang2024autonomous, jiang2024trumans}, while available HOHI data~\cite{liu2025core4d} is limited in scale and often ill-suited for learning physically grounded control policies. In addition, cooperative manipulation is much more complex than single-agent manipulation. The joint action space grows exponentially with the number of agents, leading to pronounced difficulties in optimization, convergence, and training stability. As a result, even SOTA methods often struggle to generalize to diverse interaction patterns, novel object geometries, and unseen coordination scenarios.

To overcome these hurdles, as shown in Figure.~\ref{fig:teaser}, we present \textbf{SynAgent}, a unified framework designed to bridge the chasm between single-agent motor proficiency and multi-agent collaborative synergy. Our first contribution addresses the fundamental data bottleneck through an \textbf{interaction-preserving retargeting} strategy. Direct motion transfer often suffers from severe skeletal discrepancies, so we utilize a differentiable SMPL-X proxy to bridge the morphology gap between human performers and humanoid agents. We then construct an \textit{Interact Mesh} via Delaunay tetrahedralization, which effectively encapsulates the local spatial structure between agent joints and object vertices. By minimizing the Laplacian deformation energy of interact meshes during the retargeting process, our method explicitly maintains motion semantics and contact relationships.

Second, we introduce a \textbf{Solo-to-Cooperative Agent Synergy} paradigm. We observe that cooperative manipulation can be conceptually reformulated as a single-agent control problem subject to external force disturbances. This insight allows us to leverage abundant, high-quality single-human HOI data to distill robust motion imitation priors. These individual skills are subsequently evolved into multi-agent synergy through a decentralized training scheme with shared weights, optimized via Multi-Agent Proximal Policy Optimization (MAPPO) to foster emerging collaborative behaviors. As for achieving precise execution, we develop a \textbf{trajectory-conditioned policy} as a conditional VAE, which integrates agent states, object trajectories, and geometry-aware interaction graphs. To mitigate the ill-posed trajectory-following, we employ a \textbf{Multi-Teacher Distillation} framework. By distilling the collective expertise of specialized imitation teachers into a unified student model through a progressive DAgger-style schedule, our framework produces stable, coherent, and highly generalizable multi-agent manipulation motions across diverse object geometries.

In summary, our primary contributions include:
\begin{itemize}
    \item A novel interaction-preserving retargeting method utilizing an Interact Mesh representation to explicitly maintain the semantic integrity and spatial relationships of HOHI data during motion transfer.
    \item A Solo-to-Cooperative Agent Synergy paradigm that distills scalable multi-agent cooperative behaviors from abundant single-human interaction priors via distilling and decentralized learning.
    \item A multi-teacher distillation framework that consolidates diverse motion priors into a unified trajectory-conditioned policy, enabling robust generalization across varying object geometries and trajectories.
\end{itemize}

%% file: sections/related_work.tex
\section{Related Work}

\subsection{Motion Imitation and Retargeting}

Physics-based motion imitation learns control policies to track reference motions in simulation, as demonstrated by prior work such as DeepMimic~\cite{peng2018deepmimic}and Mimickit~\cite{peng2025mimickit}, which commonly adopts policy optimization methods like PPO~\cite{schulman2017proximal}. Owing to motion imitation, robotic tasks are decomposed into two subtasks: generation and imitation. Prior work demonstrated compelling results in specialized scenarios, such as human interaction generation~\cite{yao2025physiinter,jiao2026hitmm}, sport interactions~\cite{luo2024smplolympics,wang2024skillmimic,haarnoja2024learning} and object transportation~\cite{bae2023pmp}. \cite{zhang2023simulation} achieved handless HOHI motion imitation, but we go further and use the knowledge learned from imitation to achieve more complex trajectory controlling. 

Motion retargeting aims to adapt existing motion data to new embodiments or actors. Recent works~\cite{kim2016retargeting, zhang2024skinned} have achieved impressive progress in this domain, but most of them are not directly designed for scalable multi-human interaction scenarios. For instance, Tairan et al.~\cite{he2024learning} primarily focus on retargeting motions to robotic, without addressing human–object interaction.
Meanwhile, methods such as OmniRetarget~\cite{yang2025omniretarget} and SPIDER~\cite{pan2025spider} extend retargeting to human-object and are capable of handling interactive motions. But we use interaction-preserving retargeting to refine and enhance \textbf{HOHI} data, enabling effective optimization of multi-character interactions.

\subsection{Human-Object Interaction Generation}
HOI generation gained increasing attention
as a means to synthesize realistic interactions between humans
and objects. 
Existing HOI datasets provided a strong foundation for learning interaction behaviors.
Extensive research has explored a wide spectrum,
ranging from hand–object~\cite{cha2024text2hoi,christen2024diffh2o,li2024task,ma2024diff,zhang2024artigrasp,zhang2025manidext} and static HOI generation~\cite{hou2023compositional,kim2023ncho,xie2025intertrack} to full-body and dynamic interaction synthesis~\cite{ghosh2023imos,krebs2021kit,lee2023locomotion,li2023object,razali2023action,taheri2022goal,wan2022learn,wu2022saga,xu2021d3d,zhang2022couch,yao2026hosig}.

Compared to HOI, high-quality datasets capturing 
HOHI remain scarce~\cite{loi2023machine}. CORE4D~\cite{liu2025core4d} focuses on two-person interaction synthesis. So, its interactions are constrained at a kinematic level, with limited physical fidelity. Scarce data is not the only problem, although progress in multi-agent RL has been made~\cite{vaillant2016multi}.
For example, CooHOI~\cite{gao2024coohoi} achieves collaborative multi-human interaction with objects by simplifing all objects into unified box-like proxies. Furthermore, CooHOI constructs an overly engineered reward system that is specifically tailored for box manipulation. On the contrary, our SynAgent adopts a scalable data-driven approach, which uses HOI data cleverly to compensate for the shortage of HOHI data.

%% file: sections/method.tex
\section{Method}

\noindent\textbf{Notation.} We use $\boldsymbol{p}$ and $\boldsymbol{q}$ to denote position and rotation, respectively, $\dot{\,}$ denotes the relative velocity, e.g., $\dot{\boldsymbol{p}}$ and $\dot{\boldsymbol{q}}$, and $\hat{.}$ represents ground truth.

\subsection{Interaction-Preserving Motion Retargeting}

The data processing workflow consists of four steps. We first fit a Bridge SMPL-X model to address the non-differentiable forward kinematics of the simulated agent. We then retarget by optimizing the Interact Mesh objective in Equation~\ref{eq: retargeting}, smooth retargeting results and filter low-quality data.

\begin{figure}[th]
    \centering
    \includegraphics[width=\linewidth]{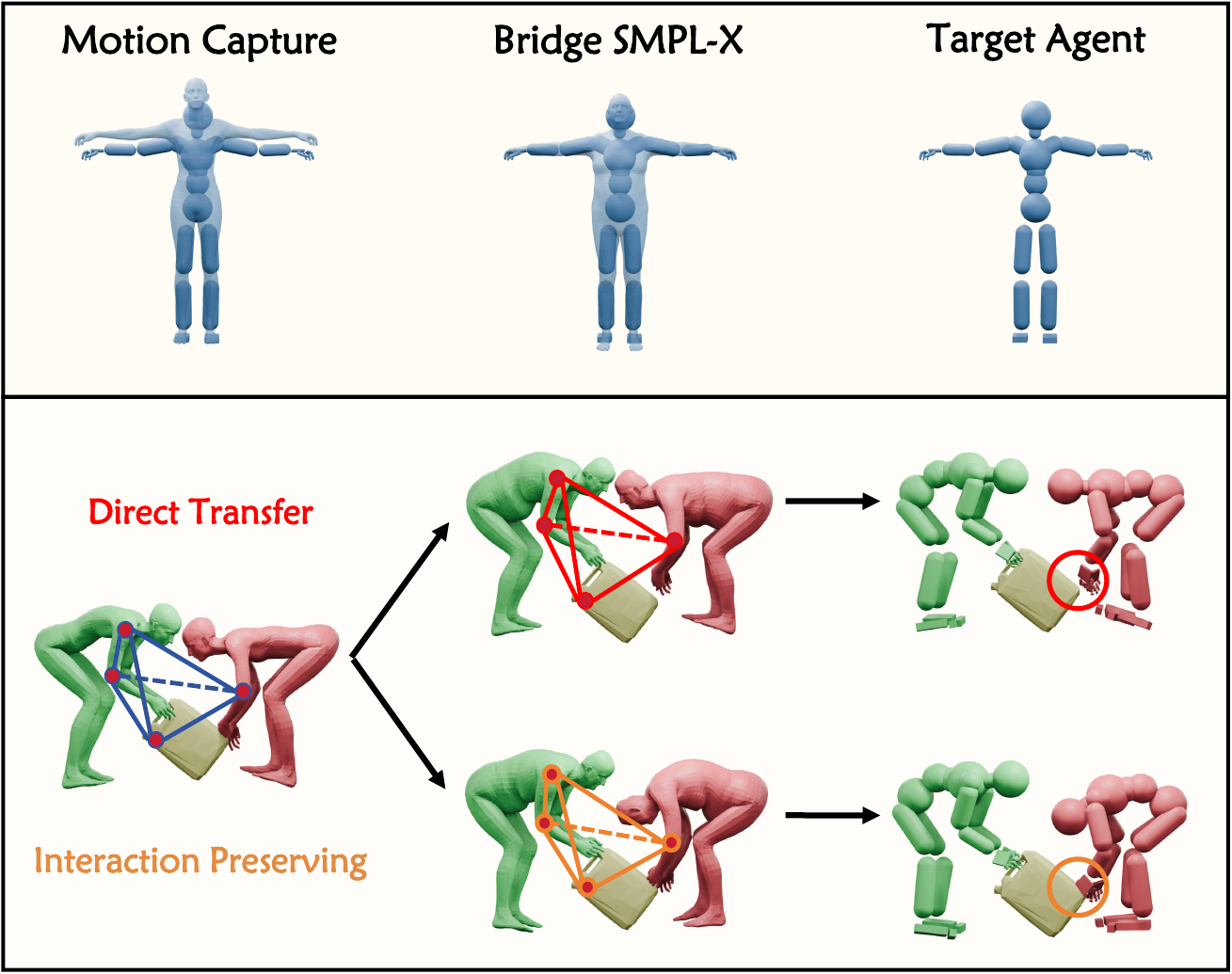}
    \caption{\textbf{Interaction-Preserving Retargeting}. (1) The upper part shows the difference between the MoCap actor and the actual agent's skeletons. (2) In the lower part, the red shows the result of directly retargeting MoCap joints onto the agent, leading to incorrect interaction relationships. The brown area shows how we construct tetrahedrons to describe the interaction relationships, making the interaction relationships unchanged after retargeting.}
    \label{fig: retarget}
\end{figure}

\textbf{Shape Fitting} Due to skeletal discrepancies between MoCap actors and simulated agents, transferring MoCap data directly often results in incorrect interaction, as highlighted by the red circles in Figure~\ref{fig: retarget}. These artifacts hinder training because agents may fail to manipulate objects even when following the reference motion. A central challenge is that the simulated agent's native forward kinematics pipeline is non-differentiable, which prevents gradients from propagating through losses defined on joint positions. We therefore perform shape fitting and get a SMPL-X~\cite{smplx} template as a differentiable proxy. By matching a SMPL-X template to the target agent's proportions, we reformulate retargeting as differentiable alignment between two SMPL-X instances and bypass the non-differentiable agent kinematics. The shape fitting objective is defined as
\begin{equation}
\begin{aligned}
\boldsymbol{\beta}_{b}^{*}=\arg \min _{\boldsymbol{\beta}_{b}}\left\|\mathcal{J}\left(\boldsymbol{\theta}_{0}, \boldsymbol{\beta}_{b}\right)-\boldsymbol{p}^{\mathrm{target}}\left(\boldsymbol{q}_{0}\right)\right\|^{2}
\end{aligned}
\end{equation}
Here, $\boldsymbol{\beta}_{b}\in\mathbb{R}^{10}$ denotes the SMPL-X shape parameters; $\mathcal{J}$ and $\boldsymbol{p}^{\mathrm{target}}$ output the SMPL-X and agent joint positions respectively; and $\boldsymbol{\theta}_{0}$ and $\boldsymbol{q}_{0}$ are their T-pose rotations. Iterative optimization yields the agent-matched shape $\boldsymbol{\beta}_{b}^{*}$, shown as the \textit{Bridge SMPL-X} in Figure~\ref{fig: retarget}.

\textbf{Retargeting} To preserve interaction semantics in complex human-object and human-object-human scenarios, we introduce an \emph{Interact Mesh} constructed via Delaunay tetrahedralization. For each interaction frame, two joints $\boldsymbol{p}_{1}, \boldsymbol{p}_{2}$ from one human, one object mesh vertex $\boldsymbol{p}_{3}$, and one joint $\boldsymbol{p}_{4}$ from the other human form a tetrahedron that captures the local interaction structure. Each tetrahedron $\tau$ composed of 4 points ${\{\boldsymbol{p}_{1},\boldsymbol{p}_{2},\boldsymbol{p}_{3},\boldsymbol{p}_{4}\}}$ is represented by the $4\times3$ Laplacian matrix $\boldsymbol{L}(\tau)=\{\boldsymbol{p}_i-\boldsymbol{p}_j\mid i,j\in\{1,2,3,4\},\,i\neq j\}$. Enforcing its invariance during retargeting preserves the relative spatial relationships among agents and objects. The retargeting objective at time step $t$ is formulated as
\begin{equation}
\label{eq: retargeting}
\begin{aligned}
{\boldsymbol{q}}_{t}^{*} =\ & \arg\min_{\boldsymbol{q}_{t}} 
\sum_{i} \big\| \boldsymbol{L}\big(\tau_{t}^{\mathrm{source}}\big) - \boldsymbol{L}\big(\tau_{t}^{\mathrm{target}}\big) \big\|^2 \\
& + \|\boldsymbol{q}_{t} - \boldsymbol{q}_{t-1}\|^2 \\
& + \max(0, \boldsymbol{q}_{\min}-\boldsymbol{q}_{t}) + \max(0, \boldsymbol{q}_{t}-\boldsymbol{q}_{\max}) \\
& + \max\!\big(0, \dot{\boldsymbol{q}}_{\min}\!\cdot\! \mathrm{d}t - (\boldsymbol{q}_{t} - \boldsymbol{q}_{t-1})\big) \\
& + \max\!\big(0, (\boldsymbol{q}_{t} - \boldsymbol{q}_{t-1}) - \dot{\boldsymbol{q}}_{\max}\!\cdot\! \mathrm{d}t \big) \\
& + \|\boldsymbol{p}_{t}^{F} - \boldsymbol{p}_{t-1}^{F}\|^2, \quad \forall\, \dot{\boldsymbol{p}}_{\mathrm{horizontal}}^{F} < 0.01~\text{m/s},
\end{aligned}
\end{equation}
where the first term enforces Laplacian consistency of the Interact Mesh $\tau_{t}^{\mathrm{source}}$ and $\tau_{t}^{\mathrm{target}}$, and the remaining terms impose temporal smoothness, joint position and velocity limits. The final term penalizes foot sliding by enforcing a stationary foot-joint position $\boldsymbol{p}^{F}$ whenever its horizontal velocity $\dot{\boldsymbol{p}}_{\mathrm{horizontal}}^{F}$ falls below 0.01~m/s. Together, these constraints yield retargeted motions that are physically plausible and faithful to the original interactions. Due to the low quality of the data and the disruption caused by retargeting, we perform additional smoothing on the retargeting data. Please refer to supplementary materials for more details.

\textbf{Filter} Some HOHI motions remain noisy after retargeting and can destabilize training. We therefore use a performance-driven \emph{train-to-filter} strategy. The policy is trained on the current dataset, and clips whose inference episode lengths fall below the mean $\sigma$ are removed before retraining. This process repeats until $\sigma$ converges, allowing the model to focus on reliable demonstrations. Note the evaluation set remains fixed, so there is no bias on evaluation.

\begin{figure*}[th]
    \centering
    \includegraphics[width=\textwidth]{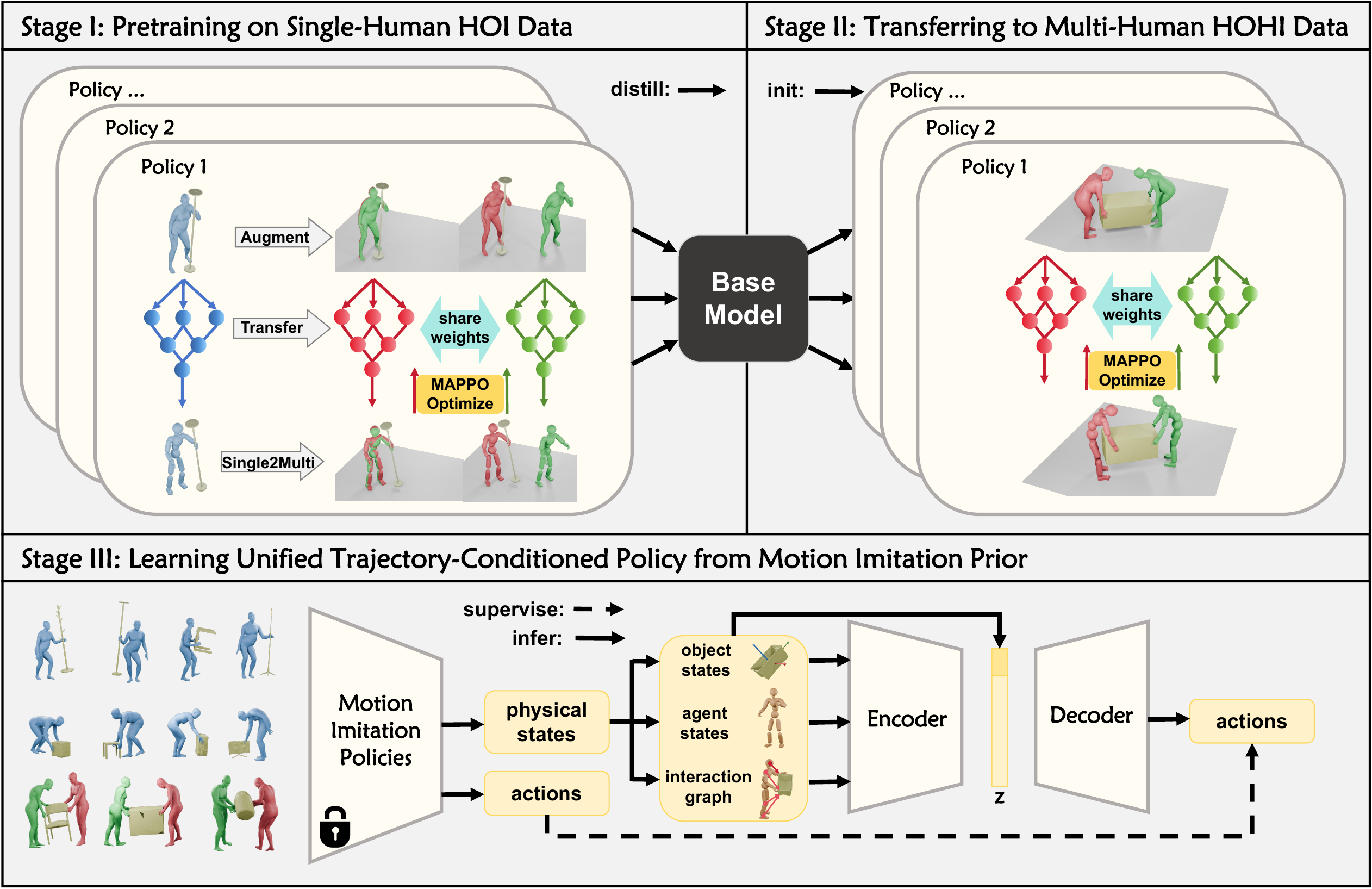}
    \caption{\textbf{Overview of SynAgent Training Pipeline.} (1) Stage I pre-trains imitation policies $\{ \boldsymbol{\pi}_i^{s} \}_{i=1}^{N}$ on single-human HOI data, then adapts them to multi-agent scenarios with MAPPO algorithm. (2) After distilling $\{ \boldsymbol{\pi}_i^{s} \}_{i=1}^{N}$ into a unified Base Model, Stage II adapts the Base Model to multi-human HOHI data and get policies $\{ \boldsymbol{\pi}_i^{m} \}_{i=1}^{M}$. (3) Stage III learns a trajectory-conditioned cVAE policy. Motion imitation policies $\{ \boldsymbol{\pi}_i^{s} \}_{i=1}^{N}$ and $\{ \boldsymbol{\pi}_i^{m} \}_{i=1}^{M}$ are used to provide physical states as refined training data, and actions for supervised learning, which stabilize and boost the model training.}
    \label{fig: pipeline}
\end{figure*}

\subsection{Single-Agent Pretraining for Cooperation}

\subsubsection{Reinforcement Learning Formulation}

Our policy follows a single-agent paradigm: at each control step, it observes the state of one agent and outputs only that agent's action. During deployment, multiple agents are independently controlled by identical copies of this shared policy, enabling decentralized cooperation.

\textbf{State} To capture motion dynamics, the state is represented by two consecutive observations $(\boldsymbol{o}_{t}, \boldsymbol{o}_{t+\Delta t})$. Each observation contains an \textbf{a}gent--\textbf{o}bject \textbf{o}bservation $\boldsymbol{o}^{\mathrm{ao}}$, an \textbf{i}nteraction \textbf{g}raph $\boldsymbol{ig}$, and a delta interaction graph $\Delta\boldsymbol{ig}=\boldsymbol{ig}-\boldsymbol{ig}_{\mathrm{ref}}$. The interaction graph represents vectors from agent joints to their nearest object-mesh vertices and thus provides geometric contact cues. The agent--object observation $\boldsymbol{o}^{\mathrm{ao}}$ is further divided into agent observations $\boldsymbol{o}^{\mathrm{a}}$ and object observations $\boldsymbol{o}^{\mathrm{o}}$. For each agent, $\boldsymbol{o}^{\mathrm{a}}$ contains joint positions $\boldsymbol{p}$, rotations $\boldsymbol{q}$, linear and angular velocities $\dot{\boldsymbol{p}}$ and $\dot{\boldsymbol{q}}$, contact indicators $\boldsymbol{c}$, and deviations from the reference motion. The object observation $\boldsymbol{o}^{\mathrm{o}}$ contains position $\boldsymbol{p}_{\mathrm{o}}$, orientation $\boldsymbol{q}_{\mathrm{o}}$, linear and angular velocities $\dot{\boldsymbol{p}}_{\mathrm{o}}$ and $\dot{\boldsymbol{q}}_{\mathrm{o}}$, and deviations from the reference object trajectory.

\textbf{Reward} The reward function is designed to encourage accurate motion imitation while maintaining physical realism during cooperative manipulation. It can be summarized as
\begin{equation}
R = e^{-\lambda_{\Delta}(\boldsymbol{\Delta} \cdot \boldsymbol{\omega})} \cdot
e^{-\lambda_c \sum \|\hat{\boldsymbol{c}} - \boldsymbol{c}\| \odot \hat{\boldsymbol{c}}} \cdot
e^{-\lambda_v \sum \|\boldsymbol{v}\| - \lambda_f \max \|\boldsymbol{f}\|},
\label{eq: reward}
\end{equation}
The first term penalizes deviations $\boldsymbol{\Delta}$ between simulated states and reference motions. Contact labels are defined by joint-to-mesh distance: distances below $0.07$ m encourage contact, those from $0.07$ m to $0.2$ m are ignored, and those above $0.2$ m penalize undesired contact. The contact term aligns simulated contacts $\boldsymbol{c}$ with reference contacts $\hat{\boldsymbol c}$, while the energy-efficiency term suppresses high-frequency joint motion $\boldsymbol{v}=\{\dot{\boldsymbol{p}}, \dot{\boldsymbol{q}}\}$ and excessive contact forces $\boldsymbol{f}$. In Eq.~\ref{eq: reward}, $\boldsymbol{\lambda}$ and $\boldsymbol{\omega}$ denote groups of hyper-parameters for different state components. For showing our method is generalizable, we use the same hyper-parameters across all objects, motions, and tasks without task-specific tuning.

\textbf{Policy \& Action} We parameterize the policy with a standard actor--critic MLP. It receives the concatenated observation $(\boldsymbol{o}_{t},\boldsymbol{o}_{t+\Delta t})$ and predicts one agent's joint PD targets $\boldsymbol{a}_{t}\in\mathbb{R}^{51\times3}$ in exponential-map form. These targets are converted into torques for the whole-body humanoid.

\subsubsection{Training Stages of Imitation Policies} 
\label{sec: stages of imitation}
Our core idea is to learn fundamental skills from simple tasks and progressively transfer them to more complex scenarios involving two agents. Throughout this process, we keep the network architecture, hyper-parameters, and reward function unchanged. 

\textbf{Stage I} In the first step of Stage~I, we partition the single-human data into $N$ subsets and train a separate policy on each subset, corresponding to the blue networks in Stage~I of Figure~\ref{fig: pipeline}. Subsequently, we perform three operations. First, we augment the single-human data by duplicating the original human. The duplicate is either placed identically to the original or randomly positioned nearby. On this augmented data, we proceed to train a multi-agent policy. For the two agents in the scene (the red and green agents in Figure~\ref{fig: pipeline}), we employ two networks with shared weights, initialized with the previously trained single-agent policy. A key design is the joint training of this policy using MAPPO algorithm, whose core objective is given by 
\begin{equation}
V_{\boldsymbol{\phi}} = \arg\min_{\boldsymbol{\phi}} \frac{1}{|\text{agent}|T}\!\sum_{\text{agent} \in \mathcal{S}}\!\sum_{t=1}^{T}\!\left(V_{\boldsymbol{\phi}}\!(\boldsymbol{o}_{t}, \boldsymbol{o}_{t+\Delta t}) - \hat{R}_{t}\right)^2
\label{eq: mappo}
\end{equation}
where $V_{\boldsymbol{\phi}}$ is the critic network, $|\text{agent}|$ is the number of agents in a simulated scene $\mathcal{S}$, and $\hat{R}_{t}$ is the real reward, i.e., $R$ in Equation~\ref{eq: reward}. This function aims to integrate information from all agents to jointly optimize the value function model, thereby helping the policy learn collaborative skills. At last, as illustrated at the bottom of Stage~I in Figure~\ref{fig: pipeline}, both agents are trained simultaneously in a shared environment. We disable collisions between agents but enable collisions between each agent and the object. This design allows us to directly leverage single-human data to train multi-agent cooperative skills without forgetting the single-agent abilities, and the agents can perceive each other's presence through the object's dynamics. Inter-agent collisions are disabled only in Stage I because the duplicated agents may overlap; all collisions are enabled during subsequent training and inference.

\textbf{Stage II} We distill the $N$ single-human policies $\{\boldsymbol{\pi}_i^s\}_{i=1}^{N}$ into an architecture-matched \emph{Base Model} using the similar procedure in the next section. This Base Model initializes subsequent training on $M$ subsets of HOHI data. Table~\ref{tab: imit abla} shows that the distilled prior substantially stabilizes and accelerates cooperation learning compared with training from scratch. The resulting policies $\{\boldsymbol{\pi}_i^s\}_{i=1}^{N}$ and $\{\boldsymbol{\pi}_i^m\}_{i=1}^{M}$ constitute the Stage III \emph{Motion Imitation Policies} and provide state--action supervision for final training.

\subsection{Generalizable Trajectory-Conditioned Policy}
\label{sec: distill}

As shown in Stage III of Figure~\ref{fig: pipeline}, our final model, \textbf{SynAgent}, is essentially a conditional Variational Autoencoder (VAE). Unlike the imitation policy described earlier, SynAgent removes all observations related to the reference state of the agent from its input. The encoder receives three streams of information: the current \emph{agent states} $\boldsymbol{o}^{\mathrm{a}}$, the \emph{interaction graph} $\boldsymbol{ig}$, and the \emph{object states} $\boldsymbol{o}^{\mathrm{o}}$ ---where the object states $\boldsymbol{o}^{\mathrm{o}}$ still include the reference object state as a control signal. The condition for our final model is defined as a \(T \times 3\) sequence of translations, representing the 3D position of the object across \(T\) timesteps. The encoder outputs a latent vector $\boldsymbol{z}$, which is then concatenated with $[\boldsymbol{o}^{\mathrm{o}},\boldsymbol{p},\boldsymbol{q},\boldsymbol{\dot{p}},\boldsymbol{\dot{q}}, \boldsymbol{ig}]$,  and passed to the decoder to generate the corresponding actions $\boldsymbol{a}$. In the following, we introduce two core designs of our training algorithm that enable effective learning of this trajectory-conditioned generative policy, which are illustrated in Algorithm~\ref{alg:distill_rl}.

\begin{algorithm}[t]
\caption{Learning from Motion Imitation Prior}
\label{alg:distill_rl}
\begin{algorithmic}[1]
\REQUIRE Motion imitation policies $\boldsymbol{\pi}^{\mathrm{imit}}$ composed of $\{ \boldsymbol{\pi}_i^{s} \}_{i=1}^{N}$ and $\{ \boldsymbol{\pi}_i^{m} \}_{i=1}^{M}$, SynAgent policy parameters $\boldsymbol{\psi}$, SynAgent value function parameters $\boldsymbol{\phi}$, DAgger hyper-parameter $\epsilon$ and $\kappa$, horizon length $H$, max episode rounds $L$ of imitation training

\FOR{episode $l = 0, 1, 2, \ldots$}
    \FOR{$t = 1$ \TO $H$}
        \STATE Sample a variable $u \sim \mathrm{Uniform}(0,1)$
        \STATE Obtain $\boldsymbol{a}_{t}^{e}$ from $\boldsymbol{\pi}^{\mathrm{imit}}(\boldsymbol{a} \mid \boldsymbol{s}_{t}^{e})$
        \STATE Obtain $\boldsymbol{a}_{t}$ from $\boldsymbol{\pi}_{\boldsymbol{\psi}}(\boldsymbol{a} \mid \boldsymbol{s}_{t}^{e})$
        \IF{$u \leq \max \left( 1-\max \left( \frac{l-\kappa}{\epsilon}, 0 \right) , 0 \right)$}
            \STATE Given $\boldsymbol{s}_{t}^{e}$, execute $\boldsymbol{a}_{t}^{e}$, observe $\boldsymbol{s}_{t+1}^{e}$
        \ELSE
            \STATE Given $\boldsymbol{s}_{t}^{e}$, execute $\boldsymbol{a}_{t}$, observe $\boldsymbol{s}_{t+1}^{e}$
        \ENDIF
        \STATE Store the transition $(\boldsymbol{s}_{t}^{e}, \boldsymbol{s}_{t+1}^{e}, \boldsymbol{a}_{t}, \boldsymbol{a}_{t}^{e})$
    \ENDFOR
    \STATE Update $\boldsymbol{\phi}$ according to Eq.~\ref{eq: mappo}
    \STATE Compute $J(\boldsymbol{\psi}) = \lVert \boldsymbol{a} - \boldsymbol{a}^{e} \rVert$
    \STATE Compute $w = \min\!\left(\max\!\left(1-\max\!\left(\frac{l-\kappa}{\epsilon},0\right),0\right),1\right)$
    \IF{$l < L$}
        \STATE Compute imitation PPO objective: $\mathcal{L}_{\mathrm{imit}}(\boldsymbol{\psi})$
        \STATE Update $\boldsymbol{\psi}$ by $\nabla_{\boldsymbol{\psi}}\left(w\mathcal{L}_{\mathrm{imit}}(\boldsymbol{\psi}) + (1-w)J(\boldsymbol{\psi})\right)$
    \ELSE
        \STATE Compute trajectory PPO objective: $\mathcal{L}_{\mathrm{traj}}(\boldsymbol{\psi})$
        \STATE Update $\boldsymbol{\psi}$ by $\nabla_{\boldsymbol{\psi}}\left(w\mathcal{L}_{\mathrm{traj}}(\boldsymbol{\psi}) + (1-w)J(\boldsymbol{\psi})\right)$
    \ENDIF
\ENDFOR
\end{algorithmic}
\end{algorithm}

\input{sections/experiment_main_tables}

\textbf{Distill from Imitation to Generation} Unlike motion imitation, trajectory-conditioned control is an ill-posed problem. There exist infinitely many possible action sequences that can move an object along a given trajectory, which often leads to training instability and convergence difficulties. To address this, we propose to learn action generation from motion imitation priors. As illustrated in Stage~III of Figure~\ref{fig: pipeline}, we first process the original reference data using our pre-trained imitation policies $\boldsymbol{\pi}^{\mathrm{imit}}$ to produce physically consistent state--action pairs: physical states $\boldsymbol{s}^{e}$ and corresponding actions $\boldsymbol{a}^{e}$. The physical states $\boldsymbol{s}^{e}$ enhance the raw motion capture data by ensuring full physical plausibility, while the actions $\boldsymbol{a}^{e}$ provide direct supervision for the generative model. This strong pairing effectively reduces the under-constrained nature of trajectory control, transforming an open-ended problem into a well-defined one with a unique solution per demonstration.

\textbf{Progressive Skill Acquisition} The complete algorithm flow is shown in Algorithm~\ref{alg:distill_rl}. To ensure stable training and enable a gradual transition from pure imitation of $\boldsymbol{\pi}^{\mathrm{imit}}$ to autonomous action generation, we introduce three key hyper-parameters to control the learning schedule. The first two, $\epsilon$ and $\kappa$, are adopted from the DAgger algorithm. When $l\leq\kappa$, only teacher actions $\boldsymbol{a}^{e}$ are executed because the untrained policy $\boldsymbol{\psi}$ would cause early termination. During $\kappa<l<\epsilon+\kappa$, we linearly reduce both the probability of sampling $\boldsymbol{\pi}^{\mathrm{imit}}$ actions and the weight of the action-imitation loss $J(\boldsymbol{\psi})$. After $\kappa+\epsilon$ episodes, SynAgent executes only its own actions and $J(\boldsymbol{\psi})$ is removed. The third hyperparameter, $L$, governs the reward composition. 
As outlined in Algorithm~\ref{alg:distill_rl}, for the first $L$ episode rounds, we employ the full imitation reward used in Stages I and II to ground the policy. Beyond $L$, we switch to a sparse reward that depends only on the object's trajectory tracking error. This shift encourages the policy to explore its own action sequences to achieve the goal, rather than slavishly mimicking the teacher's motions, thereby significantly improving generalization to novel trajectories and interaction scenarios.

%% file: sections/experiment_main_tables.tex
\begin{table*}[!t]
\centering
\footnotesize
\setlength{\tabcolsep}{4.4pt}
\begin{tabular}{cccccccccccc}
\hline
\multicolumn{2}{c}{Method}                                    & box              & bucket           & chair            & clothes stand     & table            & floor lamp        & monitor          & suitcase        & tripod           & all              \\ \hline
                                                 & Dist.    $\downarrow$     & 1.1801           & 1.382            & 1.4404           & 1.3825           & 1.2835           & 1.3837           & 1.1688           & 1.3088          & 1.3212           & 1.3074           \\
\multirow{-2}{*}{CooHOI}                         & Succ. Rate $\uparrow$ & \textbf{14.14\%} & 2.95\%           & 4.26\%           & 0.00\%           & 10.66\%          & 6.56\%           & 1.64\%           & 3.28\%          & 0.00\%           & 7.28\%           \\
\rowcolor[HTML]{EFEFEF}
\cellcolor[HTML]{EFEFEF}                         & Dist.    $\downarrow$     & \textbf{0.9985}  & \textbf{0.3871}  & \textbf{0.4099}  & \textbf{0.6524}  & \textbf{0.7488}  & \textbf{0.4271}  & \textbf{0.6009}  & \textbf{1.1005} & \textbf{0.8179}  & \textbf{0.7939}  \\
\rowcolor[HTML]{EFEFEF}
\multirow{-2}{*}{\cellcolor[HTML]{EFEFEF}SynAgent} & Succ. Rate $\uparrow$ & 11.88\%          & \textbf{51.79\%} & \textbf{39.33\%} & \textbf{16.39\%} & \textbf{12.94\%} & \textbf{37.23\%} & \textbf{25.00\%} & \textbf{5.65\%} & \textbf{15.48\%} & \textbf{19.16\%} \\ \hline
\end{tabular}
\caption{Comparison for Control. Note that each major category of objects also has various subcategories. See the supplementary materials for detailed experimental results.}
\label{tab: control}
\end{table*}

%% file: sections/experiments.tex
\section{Experiments}

\subsection{Implementation Details}

Our experiments are conducted on the OMOMO and CORE4D datasets. OMOMO provides HOI sequences, while CORE4D contains HOHI data. After automatic filtering to remove low-quality samples, we obtain 2,960 motion sequences covering 9 object categories and 25 distinct objects. Based on these data, we train 20 motion imitation models in total, including $N=17$ on OMOMO and $M=3$ on CORE4D. Each imitation model is trained on a single NVIDIA RTX 3090 GPU, while the final distillation stage for learning the generalizable trajectory-conditioned policy is performed on an NVIDIA A800 GPU.

\textbf{Comparison}  
Since no prior method is explicitly designed for HOHI imitation or trajectory-controlled cooperative humanoid manipulation, we adopt representative alternatives for comparison. For motion imitation, we use InterMimic~\cite{xu2025intermimic}, a state-of-the-art HOI imitation method, applied independently to each agent to generate actions. For trajectory-conditioned control, we compare against CooHOI~\cite{gao2024coohoi}, which supports goal-conditioned object manipulation but does not track trajectories. Both methods are evaluated on their ability to cooperatively move an object to a target position, enabling a fair comparison at the object-control level. Qualitative comparison is shown in supplementary materials. Please refer to the demo in the Supp. Mat. for more qualitative results. Meanwhile, besides the results presented in the main paper, the Supp. Mat. contains a wealth of more detailed evaluation results for reference.

\textbf{Ablation Studies}  
We perform ablation studies to examine key design choices in our framework. These include comparing centralized and decentralized policy structures for dual-agent manipulation, evaluating different architectural variants of the VAE-based trajectory-conditioned policy, and assessing the transferability of knowledge learned from single-human data to multi-human cooperation. We also ablate the proposed data refinement strategies, including interaction-preserving retargeting and automatic filtering, to quantify their effects on training stability and performance.

\textbf{Metrics}
We use task-specific metrics to evaluate both imitation and control. For motion imitation, we report the success rate (Succ. Rate), episode length (Ep. Len.), and accumulated reward (Reward). We further measure accuracy using the success-rate-weighted mean human joint error $E_h$ and object trajectory error $E_o$, where lower values indicate more accurate imitation. For trajectory-conditioned control, we report the success rate based on a final-position threshold and the average final-position error across all trials, with smaller errors indicating more precise goal-directed manipulation. Please refer to the Supp. Mat. for more details.

\subsection{Imitation and Control}

\textbf{Comparison of Imitation.} Table~\ref{tab: imit} compares InterMimic and our method under the imitation setting. Both methods are trained on the same augmented OMOMO dataset using identical data splits, where we train 17 imitation policies $\{ \boldsymbol{\pi}_i^{s} \}_{i=1}^{N}$ for our approach and 17 corresponding models for InterMimic. By the way, single-human data augmentation strategy is described in Stage I. As shown in the Table~\ref{tab: imit}, although InterMimic achieves strong performance in single-human HOI imitation, its performance degrades significantly when an additional agent is introduced as a source of physical interference, indicating limited adaptability to multi-human settings without targeted training. In contrast, our method substantially improves the imitation success rate and consistently outperforms InterMimic across all metrics. Notably, these gains are achieved without modifying the model architecture or tuning hyper-parameters, demonstrating that MAPPO-based joint training is critical for enabling robust multi-agent imitation and highlighting the superiority of our approach in cooperative scenarios.

\begin{table}[th]
\centering
\resizebox{\columnwidth}{!}{%
\begin{tabular}{cccccc}
\hline
Method        & Succ. Rate $\uparrow$      & Ep. Len. $\uparrow$        & Reward $\uparrow$          & $E_h$ $\downarrow$             & $E_o$ $\downarrow$             \\ \hline
InterMimic    & 7.26\%           & 51.9151          & 20.1757          & 1.7352          & 1.4329          \\
$\{ \boldsymbol{\pi}_i^{s} \}_{i=1}^{N}$          & \textbf{45.00\%} & \textbf{78.7416} & \textbf{25.5555} & \textbf{0.2376} & \textbf{0.2387} \\ \hline
\end{tabular}%
}
\caption{Comparison for Imitation in OMOMO. More detailed results are on supplementary materials.}
\label{tab: imit}
\end{table}

\textbf{Comparison of Control.} The closest existing baseline to our task is CooHOI, as both methods are able to cooperatively transport an object to a specified target position. As reported in Table~\ref{tab: control}, except for box-shaped objects, our method consistently outperforms CooHOI across nearly all object categories. CooHOI is target-conditioned, so we provided it with the final 2D positions of test trajectories. And for fairness, only the distance on 2D plane is calculated when calculating the metrics. Moreover, CooHOI is an AMP-like method, meaning no training set bias existed. CooHOI's drastic performance degradation stems from its inherent design: it abstracts objects into simple 8-keypoint bounding boxes and lacks dexterous hand modeling. Its performance relies on an over-engineered reward function specifically tailored for picking up the box, which inevitably limits its generalization to other objects. Consequently, it suffers from severe generalization failures when faced with non-box-like (even non-square boxes), complex geometries.

Moreover, CooHOI does not model dexterous hands, which severely limits its ability to manipulate objects that require stable grasping, such as clothes stands, monitors, and tripods, where its success rate is nearly 0.00\%. In contrast, our method demonstrates substantially stronger generalization. The VAE-based policy enables fine-grained trajectory-conditioned control rather than pure goal-conditioned behavior, while the interaction graph provides a more expressive and geometry-aware object representation. Although incorporating dexterous hands significantly increases learning difficulty, our framework successfully overcomes this challenge, allowing a single model to handle a broader range of objects that require precise and coordinated manipulation.

\subsection{Ablation and Analysis}

\textbf{Centralized or Decentralized} Before training the imitation policies on CORE4D in Stage II, we need to determine which policy architecture can better facilitate dual-agent cooperation. We therefore conduct an architectural ablation study at the Base Model stage. A straightforward approach is a centralized policy, where a single model observes both agents and directly outputs actions for both of them. In principle, such a design allows more explicit information sharing and could simplify coordination between agents. In practice, however, doubling the action space significantly increases the learning difficulty, leading to slower convergence and degraded performance. As a result in Table~\ref{tab: centralized decentralized}, we observe inferior results compared to alternative designs. We therefore adopt a decentralized policy structure, where each agent is controlled by a shared-weight policy that outputs its own actions. This design not only yields better empirical performance but also offers superior scalability, as it naturally extends to scenarios involving varying numbers of agents.

\begin{table}[th]
\centering
\resizebox{\columnwidth}{!}{%
\begin{tabular}{cccccc}
\hline
Method        & Succ. Rate $\uparrow$      & Ep. Len. $\uparrow$        & Reward $\uparrow$          & $E_h$ $\downarrow$             & $E_o$ $\downarrow$             \\ \hline
Centralized   & 12.61\%          & 77.0029          & 17.9975          & 0.8170          & 0.8184          \\
Decentralized & \textbf{19.92\%} & \textbf{84.0158} & \textbf{19.0685} & \textbf{0.5404} & \textbf{0.5568} \\ \hline
\end{tabular}%
}
\caption{Ablation studies of Centralized vs Decentralized Base Model in OMOMO. \textbf{Centralized} means one policy outputs two agents' actions, and \textbf{Decentralized} means two share-weight policies output two agents' actions separately.}
\label{tab: centralized decentralized}
\end{table}

\textbf{Solo-to-Cooperative Skill Transition} As a core contribution of our work, we need to explicitly validate its effectiveness in facilitating multi-agent skill learning. As shown by the results of \textit{w/o init} in Table~\ref{tab: imit abla}, training the multi-human imitation policies $\{ \boldsymbol{\pi}_i^{m} \}_{i=1}^{M}$ without initializing from the Base Model distilled from single-human policies $\{ \boldsymbol{\pi}_i^{s} \}_{i=1}^{N}$ leads to a sharp performance degradation. Without a strong skill prior acquired from abundant single-human data, learning complex cooperative behaviors from scratch becomes extremely difficult, and the training process often fails to converge. These results demonstrate that transiting single-agent skills is crucial for stable and effective learning in multi-agent cooperative manipulation tasks.

\begin{table}[th]
\centering
\setlength{\tabcolsep}{1pt} 

\resizebox{\columnwidth}{!}{%
\begin{tabular}{ccc ccccc}
\toprule
\multicolumn{3}{c}{Components} & \multicolumn{5}{c}{Metrics} \\
\cmidrule(r){1-3} \cmidrule(l){4-8} 
Init & Retarget & Filter & Succ. Rate $\uparrow$ & Ep. Len. $\uparrow$ & Reward $\uparrow$ & $E_h$ $\downarrow$ & $E_o$ $\downarrow$ \\ 
\midrule

-- & \checkmark & \checkmark & 3.33\% & 79.3216 & 16.3713 & 4.8528 & 4.4174 \\

\checkmark & -- & \checkmark & 2.42\% & 98.4375 & 20.2552 & 6.6776 & 6.0785 \\

\checkmark & \checkmark & -- & 1.43\% & 69.7496 & 13.7437 & 12.6062 & 7.9756 \\

\checkmark & \checkmark & \checkmark & \textbf{21.82\%} & \textbf{108.4463} & \textbf{20.7419} & \textbf{0.7406} & \textbf{0.6741} \\ 
\bottomrule
\end{tabular}%
}
\caption{Ablation Studies in CORE4D. We analyze the contribution of each component: initialization with Base Model (Init), Interaction-preserving retargeting (Retarget), and train-to-filter strategy (Filter). \checkmark indicates the component is used.}
\label{tab: imit abla}
\end{table}

\textbf{Data Refinement} Another major contribution is a comprehensive data refinement pipeline for HOHI motion data, whose effectiveness is reflected in the model performance. As shown by the results of \textit{w/o retarget} in Table~\ref{tab: imit abla}, interaction-preserving retargeting significantly improves learning by correcting motion artifacts caused by shape discrepancies between human performers and humanoid agents. However, retargeting alone is insufficient. The poor performance of the \textit{w/o filter} variant highlights that MoCap data contains a substantial number of erroneous and low-quality sequences that cannot be remedied through retargeting alone. Our automatic filtering strategy effectively removes such problematic samples, allowing the model to focus on reliable demonstrations. Together, retargeting and filtering address the fundamental data quality issues in HOHI learning, enabling stable training and substantially improved performance.

\textbf{Limitation} Despite its effectiveness, our work is still an initial step toward cooperative humanoid manipulation. The primary limitation lies in the scarcity of high-quality HOHI data, which prevents our model from scaling to achieve performance leaps. As a result, learning more diverse and complex interaction behaviors remains challenging.

%% file: sections/conclusion.tex
\section{Conclusion}
We presented SynAgent, a framework for generalizable cooperative humanoid manipulation that addresses the core challenges of data scarcity, coordination complexity, and trajectory control in multi-agent settings. By introducing an interaction-preserving retargeting strategy, we substantially improved the quality and usability of scarce HOHI data. Our single-agent pretraining paradigm demonstrates that cooperative manipulation can be effectively distilled from abundant single-human experience, enabling robust decentralized coordination without specialized multi-agent supervision. Furthermore, by distilling motion imitation priors into a trajectory-conditioned generative policy, SynAgent achieves precise and stable object-level control while maintaining strong generalization across object geometries and agent counts. Together, these components establish a scalable pathway from individual manipulation skills to coordinated multi-humanoid behaviors, offering a practical foundation for future research in embodied intelligence, physics-based animation, and multi-agent robotic interaction.

%% file: sections/supplementary_material.tex
\section{Supplementary Material}

\subsection{Data Preprocessing}

\textbf{Smooth} Although we incorporate smoothness terms in the retargeting objective, the optimized motion sequence may still exhibit jitter. To address this, we apply a post-processing step to the root translation and joint rotations. For the root trajectory $\boldsymbol{\rho}$, we employ Sobolev norm regularization. By solving linear equations $(\boldsymbol{I} + \alpha (\boldsymbol{D}^2)^{T} \boldsymbol{D}^2) \boldsymbol{\rho}^{*} = \boldsymbol{\rho}$, we get smoothed root trajectory  $\boldsymbol{\rho}^{*}$, where $\alpha$ is regularization parameter and $\boldsymbol{D}^2$ is second-order difference operator matrix. For joint rotations, we leverage the pre-trained SmoothNet\cite{Zeng2021SmoothNetAP} filter, a lightweight temporal network designed specifically for motion de-jittering. It processes the rotation sequence in a sliding-window manner, effectively suppressing jitter while preserving the essential kinematic posture. This decoupled approach ensures both positional and orientational smoothness in the final retargeted motion.

\textbf{Filter} We initially organized the real-world motion capture data into 7 distinct subsets based on object categories to facilitate specialized training. However, as noted above, our HOHI imitation policies incorporate only 3 policies $\{ \boldsymbol{\pi}_i^{m} \}_{i=1}^{3}$. This reduction is necessitated by the critically low quality of the raw CORE4D data, which frequently exhibit severe artifacts such as abrupt trajectory discontinuities, skeletal distortions, and physically implausible contacts. These fundamental defects rendered training non-convergent for several categories, including boards and sticks, and could not be rectified through retargeting or post-optimization. Consequently, we were forced to discard the compromised subsets entirely. 

\subsection{Final Dataset}

\begin{figure}[th]
    \centering
    \includegraphics[width=\columnwidth]{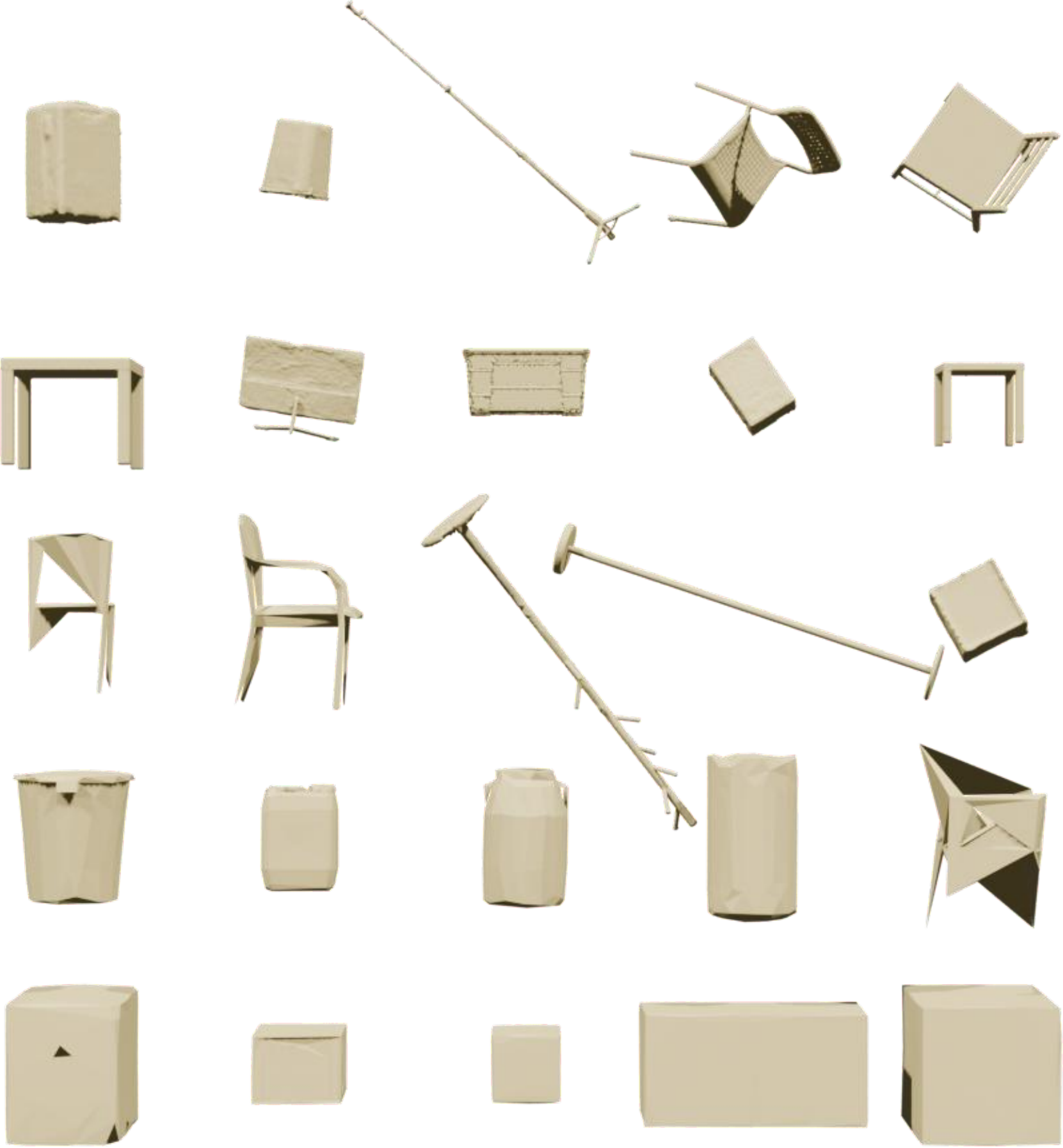}
    \caption{\textbf{Overview of 25 Objects.} Our model can ultimately cover these 25 objects.}
    \label{fig: objects}
\end{figure}

Ultimately, these sequences from both the HOI and HOHI datasets undergo refinement via our imitation policies to generate the physically consistent state-action pairs for training the final Trajectory-Conditioned Policy. By aggregating these resources, we construct a comprehensive training corpus consisting of 20 distinct subsets, comprising 17 derived from single-human data and 3 from multi-human interactions, yielding a total of 2,960 physically validated motion sequences. This diverse dataset encompasses 9 major object categories and 25 unique object geometries, the full variety of which is visualized in Figure~\ref{fig: objects}. To promote reproducibility and facilitate further research in the community, we will release both the curated training dataset and our complete source code upon the acceptance of this paper.

\subsection{Training Split}

\textbf{Training on OMOMO}
The OMOMO dataset serves as our primary source of single-human motion capture data, comprising approximately 10 hours of human-object interactions. To accommodate morphological variations, we partitioned the raw data into 17 distinct subsets based on the body shapes of the actors, yielding a total of 6,435 motion sequences. We initially trained separate imitation policies on these subsets and subsequently extended the training to dual-agent scenarios through data augmentation, resulting in a collection of expert policies denoted as $\{ \boldsymbol{\pi}_i^{s} \}_{i=1}^{17}$. A critical component of our pipeline is the refinement of this data using the trained policies. Since the original OMOMO dataset lacks dexterous hand articulation and contains physically inconsistent artifacts, we employed a physics-based re-tracking procedure to validate the motions. By retaining only the sequences where $\{ \boldsymbol{\pi}_i^{s} \}_{i=1}^{17}$ successfully completed the imitation, we curated a high-quality dataset of 2,888 physically plausible motions. Finally, this refined dataset was utilized to distill $\{ \boldsymbol{\pi}_i^{s} \}_{i=1}^{17}$ into a unified Base Model, providing a robust initialization for subsequent multi-agent learning.

\textbf{Training on CORE4D}
In the second stage, we leverage the CORE4D dataset, a large-scale benchmark for human-object-human interaction (HOHI) that nominally comprises 11,000 collaboration sequences spanning 3,000 real and virtual object shapes. However, upon closer inspection, we observed that the synthetically generated portion of the dataset exhibits low physical fidelity and contains significant artifacts, rendering it unsuitable for training robust physics-based control policies. Consequently, we excluded these synthetic samples entirely and restricted our training set exclusively to the real-world motion capture segment. Following a data processing, we distilled the dataset down to 961 sequences, which serve as the initial training data for our multi-agent cooperative policies.

\subsection{More Experiment Results}

\subsubsection{Metrics}

We design task-specific evaluation metrics for imitation and control, respectively. For motion imitation, we report five metrics. The inference success rate (Succ. Rate) measures the probability that a policy can successfully complete an entire motion sequence when evaluated over $N$ demonstrations, where higher values indicate better robustness. Episode length (Ep. Len.) reports the average duration for which imitation remains successful before termination,
with longer episode lengths reflecting more stable tracking. The accumulated reward (Reward) evaluates imitation quality, where higher rewards correspond to closer adherence to the reference motion. 

To directly quantify geometric accuracy, we further compute the mean human joint error $E_h$ over only successful episodes, weighted by the inference success rate, such that lower values indicate more accurate human motion reproduction. Similarly, the object error $E_o$  is defined as the weighted mean trajectory error between the manipulated object and the reference object motion, where smaller values correspond to more precise object-level imitation. For trajectory-conditioned control, we adopt two complementary metrics. The success rate (Succ. Rate) is defined by a minimum distance threshold, where a trial is considered successful if the final object position lies within this threshold of the target. In addition, we report the average distance between the final object position and the target position across all trials, with smaller distances indicating more accurate goal-directed manipulation.

\begin{table}[t]
\centering
\setlength{\tabcolsep}{2pt} 
\resizebox{\columnwidth}{!}{%
\begin{tabular}{ccccccc}
\hline
   & Percent. & Succ. Rate $\uparrow$       & Ep. Len. $\uparrow$         & Reward $\uparrow$          & $E_h$ $\downarrow$             & $E_o$ $\downarrow$      \\ \hline
sub1  & 14.03\%    & 1.40\%     & 52.5902  & 9.5020  & 9.2286  & 7.83576 \\
sub2  & 1.31\%     & 10.34\%    & 65.5994  & 22.4952 & 0.9107  & 0.9204  \\
sub3  & 3.03\%     & 0.50\%     & 41.7012  & 7.2260  & 32.5400 & 22.1000 \\
sub4   & 3.80\%     & 0.40\%     & 51.5622  & 9.8790  & 21.3250 & 17.2000 \\
sub5  & 4.21\%     & 8.24\%     & 81.7084  & 18.1700 & 1.4256  & 0.4110  \\
sub6  & 5.93\%     & 3.31\%     & 56.7799  & 11.7300 & 3.0333  & 1.3909  \\
sub7  & 5.89\%     & 11.03\%    & 66.1516  & 15.7531 & 1.2155  & 1.0591  \\
sub8  & 5.16\%     & 9.06\%     & 56.6106  & 14.5854 & 1.4330  & 1.5615  \\
sub9  & 6.79\%     & 0.00\%     & 33.8088  & 3.8940  & $\infty$       & $\infty$        \\
sub10 & 9.41\%     & 15.06\%    & 53.3284  & 7.4360  & 0.9364  & 0.8907  \\
sub11 & 2.03\%     & 37.78\%    & 68.4531  & 21.4116 & 0.2622  & 0.0593  \\
sub12 & 4.48\%     & 2.02\%     & 46.4172  & 10.3300 & 6.8750  & 5.7850  \\
sub13 & 6.33\%     & 8.81\%     & 42.7553  & 6.8939  & 1.4352  & 1.4159  \\
sub14 & 7.02\%     & 4.73\%     & 52.1470  & 8.9440  & 2.6340  & 2.3915  \\
sub15 & 8.87\%     & 16.50\%    & 46.0051  & 7.6330  & 0.7909  & 0.6952  \\
sub16 & 5.39\%     & 5.04\%     & 53.3982  & 12.9800 & 3.0120  & 2.2680  \\
sub17 & 6.29\%     & 5.28\%     & 43.2625  & 6.3382  & 2.2377  & 3.0623  \\ \hline
all   & 100\%      & 7.26\%     & 51.9151  & 20.1757 & 1.7352  & 1.4329  \\ \hline
\end{tabular}%
}
\caption{Evaluation Results of InterMimic in OMOMO. Note that these are evaluation results for OMOMO in a two-agent scenario after it has been expanded to two-agent mode.}
\label{tab: imit intermimic}
\end{table}

\begin{table}[t]
\centering
\setlength{\tabcolsep}{2pt} 
\resizebox{\columnwidth}{!}{%
\begin{tabular}{ccccccc}
\hline
   & Percent. & Succ. Rate $\uparrow$       & Ep. Len. $\uparrow$         & Reward $\uparrow$          & $E_h$ $\downarrow$             & $E_o$ $\downarrow$      \\ \hline
sub1  & 14.03\%    & 29.57\%    & 82.4458  & 24.7642 & 0.3771 & 0.3923 \\
sub2  & 1.31\%     & 79.31\%    & 116.6124 & 41.7244 & 0.1111 & 0.1265 \\
sub3  & 3.03\%     & 59.20\%    & 84.7115  & 18.8653 & 0.2049 & 0.1995 \\
sub4  & 3.80\%     & 44.44\%    & 83.0679  & 23.3746 & 0.2664 & 0.2459 \\
sub5  & 4.21\%     & 34.41\%    & 93.8277  & 30.4317 & 0.3173 & 0.2647 \\
sub6  & 5.93\%     & 11.45\%    & 76.9975  & 20.1347 & 0.8044 & 0.7205 \\
sub7  & 5.89\%     & 50.77\%    & 93.6577  & 32.2836 & 0.2017 & 0.224  \\
sub8  & 5.16\%     & 60.82\%    & 103.3444 & 34.2238 & 0.1771 & 0.1771 \\
sub9  & 6.79\%     & 40.00\%    & 48.7679  & 9.0562  & 0.2483 & 0.2633 \\
sub10 & 9.41\%     & 60.58\%    & 70.589   & 24.7642 & 0.2212 & 0.2136 \\
sub11 & 2.03\%     & 80.00\%    & 95.9667  & 41.7244 & 0.1070 & 0.0616 \\
sub12 & 4.48\%     & 56.57\%    & 87.9563  & 18.8653 & 0.1792 & 0.1944 \\
sub13 & 6.33\%     & 43.81\%    & 68.8254  & 23.3746 & 0.2627 & 0.3186 \\
sub14 & 7.02\%     & 57.85\%    & 95.0348  & 30.4317 & 0.1758 & 0.1801 \\
sub15 & 8.87\%     & 44.05\%    & 59.5867  & 20.1347 & 0.2556 & 0.2427 \\
sub16 & 5.39\%     & 40.06\%    & 79.4779  & 32.2836 & 0.2569 & 0.2683 \\
sub17 & 6.29\%     & 41.01\%    & 65.4814  & 34.2238 & 0.2772 & 0.3299 \\ \hline
all   & 100\%      & 45.00\%    & 78.7416  & 25.5555 & 0.2376 & 0.2387 \\ \hline
\end{tabular}%
}
\caption{Evaluation Results of $\{ \boldsymbol{\pi}_i^{s} \}_{i=1}^{N}$ in OMOMO. The same data as Table~\ref{tab: imit intermimic} is used, and the results are compared with those in Table 4. "Ours" refers to imitation policies $\{ \boldsymbol{\pi}_i^{s} \}_{i=1}^{N}$.}
\label{tab: imit ours}
\end{table}

\textbf{Imitation on OMOMO} A comparative analysis of Table~\ref{tab: imit intermimic} and Table~\ref{tab: imit ours} highlights a critical finding: proficiency in single-agent HOI imitation is insufficient for direct application to multi-agent collaborative tasks. As evidenced in Table~\ref{tab: imit intermimic}, although the baseline method (InterMimic) achieves strong results on standard single-human OMOMO sequences, its performance degrades precipitously when applied to our augmented dual-agent scenarios. This decline is primarily attributed to the inability of the policies to accommodate the dynamic perturbations introduced by a second interacting agent. In contrast, Table~\ref{tab: imit ours} demonstrates the superior robustness of our approach. By leveraging MAPPO-based joint training, our imitation policies effectively learn to mitigate external disturbances, resulting in substantially improved resistance to physical interference. Consequently, our method achieves comprehensive improvements across nearly all evaluated subsets, reaffirming the effectiveness of our framework in bridging the gap between isolated skill acquisition and resilient collaborative execution.

\begin{table}[t]
\centering
\setlength{\tabcolsep}{2pt} 
\resizebox{\columnwidth}{!}{%
\begin{tabular}{cccccc}
\hline
   & Succ. Rate $\uparrow$       & Ep. Len. $\uparrow$         & Reward $\uparrow$          & $E_h$ $\downarrow$             & $E_o$ $\downarrow$      \\ \hline
box    & 4/272=1.47\%   & 62.7061           & 12.8191          & 9.6054          & 2.7551          \\
bucket & 5/250=2.00\%   & 78.3046           & 14.3882          & 8.9950          & 6.8500          \\
chair  & 2/248=0.81\%   & 68.8705           & 14.1121          & 19.5432         & 14.8395         \\
all    & 11/770=1.43\%  & 69.7496           & 13.7437          & 12.6062         & 7.9756          \\ \hline
box    & 9/165=5.45\%   & 109.9828          & \textbf{22.3496} & 2.7302          & 1.9137          \\
bucket & 28/250=11.20\% & 118.1502          & 20.5744          & 1.4232          & 1.2500          \\
chair  & 6/162=3.70\%   & 82.0555           & 16.9745          & 4.3027          & 4.7432          \\
all    & 43/577=7.45\%  & 102.7352          & 19.9511          & 2.0912          & 1.8778          \\ \hline
box    & 29/102=28.43\% & \textbf{110.9783} & 21.8224          & \textbf{0.5388} & \textbf{0.4618} \\
bucket & 28/84=33.33\%  & \textbf{141.04}   & \textbf{23.8459} & \textbf{0.4875} & \textbf{0.4347} \\
chair  & 15/144=10.42\% & \textbf{87.6451}  & \textbf{18.1666} & \textbf{1.6218} & \textbf{1.5854} \\
all    & 72/330=21.82\% & \textbf{108.4463} & \textbf{20.7419} & \textbf{0.7406} & \textbf{0.6741} \\ \hline
\end{tabular}%
}
\caption{Ablation Study of Train-to-Filter in CORE4D. From top to bottom, these are: no filter, first filter and second filter (i.e., $\{ \boldsymbol{\pi}_i^{m} \}_{i=1}^{M}$).}
\label{tab: imit filter}
\end{table}

\begin{table}[t]
\centering
\setlength{\tabcolsep}{2pt} 
\resizebox{\columnwidth}{!}{%
\begin{tabular}{cccccc}
\hline
   & Succ. Rate $\uparrow$       & Ep. Len. $\uparrow$         & Reward $\uparrow$          & $E_h$ $\downarrow$             & $E_o$ $\downarrow$      \\ \hline
box    & 2/102=1.96\%   & 84.2532           & 17.9991          & 7.8163          & 6.6989          \\
bucket & 7/84=8.33\%    & 91.0018           & 18.2712          & 1.9507          & 1.7394          \\
chair  & 2/144=1.39\%   & 69.0171           & 14.1104          & 12.1582         & 11.8848         \\
all    & 11/330=3.33\%  & 79.3216           & 16.3713          & 4.8528          & 4.4174          \\ \hline
box    & 2/102=1.96\%   & 109.8439          & \textbf{23.9613} & 7.9336          & 2.5408          \\
bucket & 6/84=7.14\%    & 105.8384          & 20.0035          & 2.6526          & 1.7450          \\
chair  & 0/144=0.00\%   & 86.0424           & 17.7771          & $\infty$           & $\infty$               \\
all    & 8/330=2.42\%   & 98.4375           & 20.2552          & 6.6776          & 6.0785          \\ \hline
box    & 29/102=28.43\% & \textbf{110.9783} & 21.8224          & \textbf{0.5388} & \textbf{0.4618} \\
bucket & 28/84=33.33\%  & \textbf{141.04}   & \textbf{23.8459} & \textbf{0.4875} & \textbf{0.4347} \\
chair  & 15/144=10.42\% & \textbf{87.6451}  & \textbf{18.1666} & \textbf{1.6218} & \textbf{1.5854} \\
all    & 72/330=21.82\% & \textbf{108.4463} & \textbf{20.7419} & \textbf{0.7406} & \textbf{0.6741} \\ \hline
\end{tabular}%
}
\caption{Ablation Study in CORE4D. From top to bottom, these are: no initilization with Base Model, no interaction-preserving retargeting and Ours (i.e., $\{ \boldsymbol{\pi}_i^{m} \}_{i=1}^{M}$).}
\label{tab: imit init retarget}
\end{table}

\textbf{Imitation on CORE4D} As demonstrated in Table~\ref{tab: imit filter}, we validated the effectiveness of our data refinement pipeline by applying the Train-to-Filter strategy over two iterations. A consistent trend is observed across these iterations, where all performance metrics improve following each filtration pass. Crucially, the Succ. Rate metric reveals a dual improvement: not only does the percentage of successful episodes increase, but the number of successfully learned motions also rises. This phenomenon indicates that low-quality data acts as a source of interference during training. By purging these physically inconsistent samples, the model gains the capacity to master complex motions that were previously unlearnable, thereby underscoring the value of our filtering approach. Furthermore, Table~\ref{tab: imit init retarget} provides a more granular breakdown of our ablation studies, where the performance trends across subsets align with the aggregated results, confirming the consistency and robustness of our findings.

\subsubsection{Motion Control} Complementing the aggregated results presented in the main paper, Table~\ref{tab: control detail} provides a granular performance breakdown for each of the 25 individual object geometries. Consistent with the trends observed in Table~\ref{tab: control}, our framework demonstrates superior control capabilities and robust geometric generalization, significantly outperforming the CooHOI baseline, even on the box-transport tasks for which CooHOI was specifically optimized.

\makeatletter
\setlength{\@fptop}{0pt}
\makeatother
\begin{table}[!t]
\centering
\footnotesize
\setlength{\tabcolsep}{3pt}
\begin{tabular*}{\columnwidth}{@{\extracolsep{\fill}}ccccc@{}}
\hline
\multirow{2}{*}{Object} & \multicolumn{2}{c}{CooHOI}        & \multicolumn{2}{c}{SynAgent}        \\ \cline{2-5}
                        & Dist.           & Succ. Rate      & Dist.           & Succ. Rate        \\ \hline
box001                  & 1.3959          & 22.95\%         & \textbf{0.3753} & \textbf{40.00\%}  \\
box021                  & 0.9676          & 16.39\%         & \textbf{0.411}  & \textbf{55.56\%}  \\
box023                  & 1.1619          & 1.64\%          & \textbf{0.2019} & \textbf{100.00\%} \\
box024                  & 1.1487          & 26.23\%         & \textbf{0.2965} & \textbf{71.43\%}  \\
box025                  & 0.968           & 37.70\%         & \textbf{0.3851} & \textbf{42.86\%}  \\
bucket007               & 1.2274          & 6.56\%          & \textbf{0.2817} & \textbf{85.71\%}  \\
bucket008               & 1.2214          & 3.28\%          & \textbf{0.2373} & \textbf{100.00\%} \\
bucket009               & \textbf{1.2589} & \textbf{1.64\%} & 1.3359          & 0.00\%            \\
bucket010               & 1.9481          & 1.64\%          & \textbf{0.5097} & \textbf{40.00\%}  \\
chair006                & 1.0154          & 6.56\%          & \textbf{0.6434} & \textbf{25.00\%}  \\
chair020                & 1.2553          & 1.64\%          & \textbf{0.2614} & \textbf{66.67\%}  \\
chair022                & 1.1807          & 3.28\%          & \textbf{0.2591} & \textbf{100.00\%} \\
clothesstand            & 1.3825          & 0.00\%          & \textbf{0.6524} & \textbf{16.39\%}  \\
floorlamp               & 1.3837          & 6.56\%          & \textbf{0.4271} & \textbf{37.23\%}  \\
largebox                & 1.1982          & 4.92\%          & \textbf{1.0594} & \textbf{11.64\%}  \\
largetable              & 1.3339          & \textbf{18.03\%}& \textbf{0.7253} & 13.60\%           \\
monitor                 & 1.1688          & 1.64\%          & \textbf{0.6009} & \textbf{25.00\%}  \\
plasticbox              & 1.3409          & 1.64\%          & \textbf{0.8669} & \textbf{7.79\%}   \\
smallbox                & 1.2598          & 1.64\%          & \textbf{1.0379} & \textbf{10.16\%}  \\
smalltable              & 1.2330          & 3.28\%          & \textbf{0.7812} & \textbf{12.05\%}  \\
suitcase                & 1.3088          & 3.28\%          & \textbf{1.1005} & \textbf{5.65\%}   \\
trashcan                & 1.2545          & 1.64\%          & \textbf{0.3783} & \textbf{50.33\%}  \\
tripod                  & 1.3212          & 0.00\%          & \textbf{0.8179} & \textbf{15.48\%}  \\
whitechair              & 2.5922          & 0.00\%          & \textbf{0.3356} & \textbf{54.20\%}  \\
woodchair               & 1.1585          & 9.84\%          & \textbf{0.4447} & \textbf{28.07\%}  \\ \hline
all                     & 1.3074          & 7.28\%          & \textbf{0.7939} & \textbf{19.16\%}  \\ \hline
\end{tabular*}
\caption{Comparison for Control in OMOMO and CORE4D. A more detailed version of Table~\ref{tab: control}, categorizes objects in a more specific way.}
\label{tab: control detail}
\end{table}

\clearpage

\begin{figure*}[hbp]
    \centering
    \includegraphics[width=0.97\textwidth]{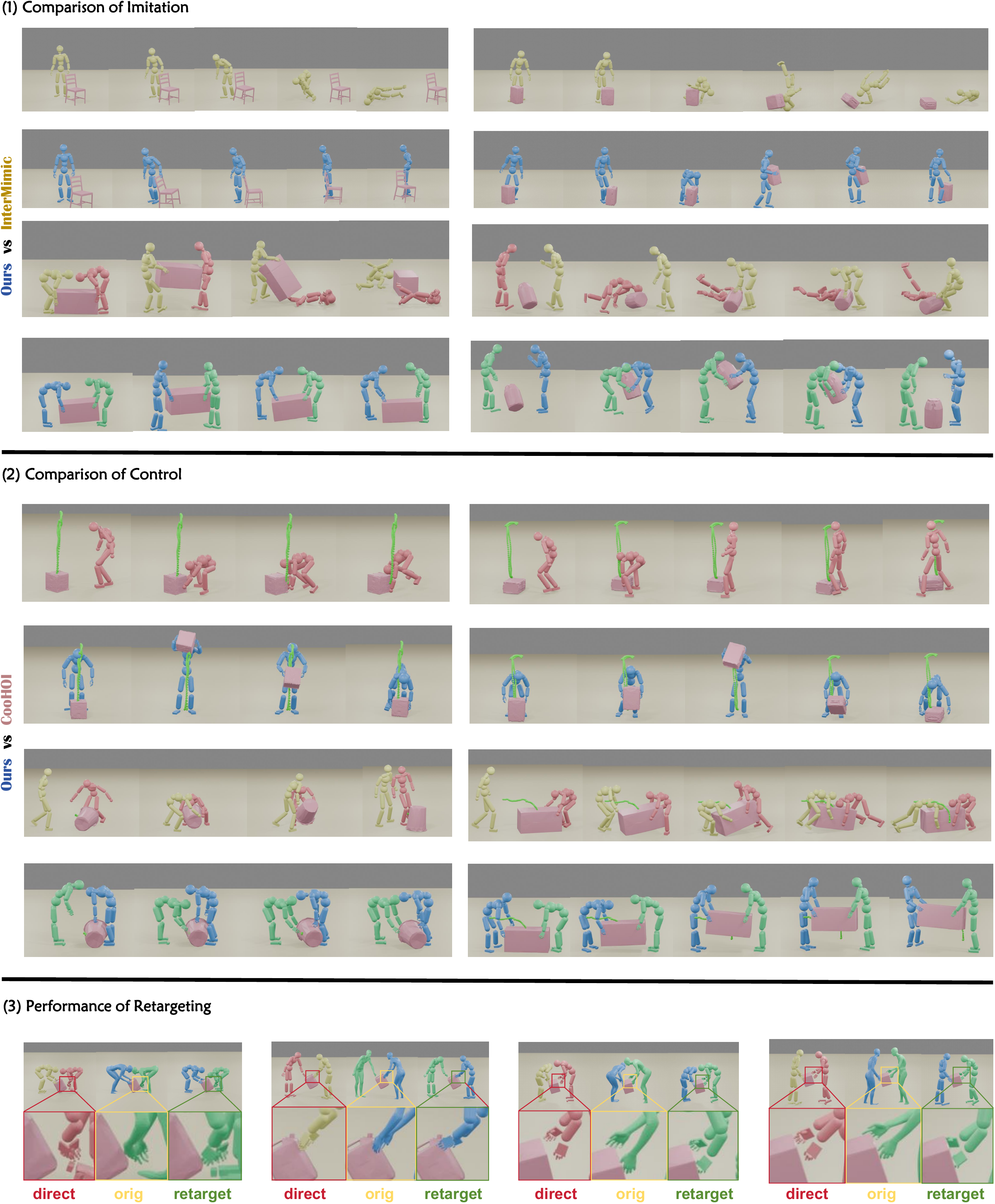}
    \caption{\textbf{Qualitative Results.} In the comparison between Ours and existing comparable baselines, the blue and green agents are the test results from Ours. In \textit{Comparison of Control}, the green ball represents the trajectory control signal. In \textit{Performance of Retargeting}, “direct" indicates that MoCap data is directly transferred to the agent, “orig" represents the raw MoCap data, and “retarget" represents the effect of using our Interaction-preserving retargeting.}
    \label{fig: qualitative}
\end{figure*}